\documentclass[conference]{IEEEtran}
\IEEEoverridecommandlockouts
\usepackage{cite}
\usepackage{amsmath,amssymb,amsfonts}
\usepackage{booktabs}
\usepackage{url}
\usepackage{algorithmic}
\usepackage{graphicx}
\usepackage{textcomp}
\usepackage{stfloats}  % allows bottom placement of figure*
\usepackage[table,xcdraw]{xcolor}
\usepackage{fancybox}
\usepackage[most]{tcolorbox}
\usepackage{todonotes}
\usepackage{xcolor}
\def\BibTeX{{\rm B\kern-.05em{\sc i\kern-.025em b}\kern-.08em
    T\kern-.1667em\lower.7ex\hbox{E}\kern-.125emX}}

\usepackage[acronym, shortcuts, nohypertypes={acronym}]{glossaries}
\newacronym{AI}{AI}{Artificial Intelligence}
\newacronym{AC}{AC}{Affective Computing}

\newacronym{ViT}{ViT}{Vision Transformer}
\newacronym{DNN}{DNN}{deep neural networks}
\newacronym{CNN}{CNN}{Convolution Neural Network}
\newacronym{FM}{FM}{Foundation Model}
\newacronym{VLM}{VLM}{Vision-Language Model}
\newacronym{SLM}{SLM}{Small-Language Model}

\newacronym{FER}{FER}{Facial Emotion Recognition}
\newacronym{UAR}{UAR}{Unweighted Average Recall}
\begin{document}

\title{Reading Smiles: Proxy Bias in Foundation Models for Facial Emotion Recognition}

\author{
    \textit{Iosif Tsangko}$^{1,2}$, \textit{Andreas Triantafyllopoulos}$^{1,2}$, \textit{Adem Abdelmoula}$^1$, \\ \textit{Adria Mallol-Ragolta}$^{1,2}$, \textit{Björn W. Schuller}$^{1,2,3,4}$ \\
    \\
    $^1$CHI -- Chair of Health Informatics, MRI, Technical University of Munich, Germany \\
    $^2$MCML -- Munich Center for Machine Learning, Munich, Germany \\
    $^3$GLAM -- Group on Language, Audio, \& Music, Imperial College London, UK \\
    $^4$MDSI -- Munich Data Science Institute, Munich, Germany \\
    
    \texttt{iosif.tsangko@tum.de}}

\maketitle
\begin{abstract}
\acp{FM} are rapidly transforming \ac{AC}, with \acp{VLM} now capable of recognising emotions in zero-shot settings. This paper probes a critical but underexplored question: \emph{what visual cues do these models rely on to infer affect, and are these cues psychologically grounded or superficially learnt}? We benchmark  varying scale \acp{VLM} on a teeth-annotated subset of AffectNet dataset and find consistent performance shifts depending on the presence of visible teeth. Through structured introspection of --the best-performing model, i.e., GPT-4o, we show that facial attributes like \emph{eyebrow position} drive much of its affective reasoning, revealing a high degree of internal consistency in its valence–arousal predictions. These patterns highlight the emergent nature of \acp{FM} behaviour, but also reveal risks: shortcut learning, bias, and fairness issues—especially in sensitive domains like mental health and education.
\end{abstract}


% \textcolor{red}{Hints:
% \begin{itemize}
% \end{itemize}

% \textcolor{orange}{
% \begin{itemize}
%   \item \textbf{Conference:} ACII 2025, Canberra, Australia, 8--11 Oct 2025
%   \item \textbf{Proceedings:} Published in IEEE Xplore
%   \item \textbf{Submission:} Double-blind, via EasyChair, max 7 pages + 1 page Ethical Impact Statement (references unlimited)
%   \item \textbf{Ethical Impact Statement:} Mandatory and reviewed; up to 1 extra page before references
%   \item \textbf{Paper Length:} 
%     \begin{itemize}
%       \item Guidelines: \url{https://acii-conf.net/2025/submission-guidelines/}
%       \item Main content: up to 7 pages
%       \item Ethical Impact Statement: up to 1 extra page
%       \item References: no page limit
%     \end{itemize}
%   \item \textbf{Preprint Policy:} arXiv allowed; ResearchGate/Academia/paid-access not allowed
% \end{itemize}}

\section{Introduction}
Understanding and interpreting human emotions is fundamental to social interaction. From early developmental cues in infants, to high-stakes decision-making in adults, facial expressions serve as a primary channel for conveying affect. Affective Computing (\ac{AC}) -the interdisciplinary field that enables machines to recognise and respond to emotional states- has evolved dramatically in recent years, transitioning from rule-based methods to powerful deep learning models~\cite{schuller18-SER, calvo10-ADA}. Within this domain, \ac{FER} plays a pivotal role, with applications in mental health, education, human–robot interaction, and automotive safety~\cite{huang19-FER}.

While modern \ac{FER} systems are often trained on large annotated datasets, the recent rise of Vision-Language Models (\acp{VLM}) and other multimodal Foundation Models (\acp{FM}) is reshaping the landscape of automatic recognition (and synthesis) of emotions~\cite{hosseini25-FOF, gao24-MLP}. \acp{VLM} are not explicitly trained for emotion classification, yet increasingly show strong zero-shot competence in affective tasks. These models are being rapidly adopted across consumer applications, mobile platforms, and edge \ac{AI} deployment~\cite{zeng23-LLMR, rahimi25-UVP, lee25-ETP, alsaad24-MLL}.

However, this rapid adoption has raised serious concerns around fairness, interpretability, and reliability~\cite{ma25-SAS}. A longstanding challenge in \ac{AI} is the tendency of models to learn proxy features-visual patterns that correlate with target labels but lack causal or semantic meaning~\cite{chen21-UAM}. This phenomenon, known as the \emph{Clever Hans effect}~\cite{kauffmann20-TCH, steinmann24-NSC, kauffmann25-EAR}, refers to systems that `appear intelligent' but in fact exploit spurious correlations in the data. Cues like teeth visibility, head pose, or eyebrow shape may function as such shortcuts, exploiting superficial correlations in the data rather than capturing genuine emotional semantics~\cite{mirabetherranz24-ADV}.

Indeed, psychophysical studies have shown that humans rely on such cues. For instance, open-mouth smiles significantly increase perceived valence and arousal, modulating both attentional and affective responses in observers~\cite{blanco17-DLA, crager16-STS}. These findings are consistent with early visual prioritisation effects demonstrated in event-related potential studies \cite{brunet25-EOC, calvo12-PCA}, where visible teeth were shown to enhance neural responses independently of emotional valence. Moreover, recent work~\cite{nomiya25-AAI} shows that these perceptual biases can influence alignment behaviour in interactive settings. Yet, despite this psychological grounding, research remains underexplored within \acp{FM}.

This paper addresses that gap by extending recent evaluation attempts, such as in~\cite{schuller24-ACH, amin23-WAC}, by conducting a comprehensive benchmark of various \acp{VLM} and comparing them against a supervised baseline.
Our best-performing model -GPT-4o- is additionally prompted to provide predictions of interpretable features.
We investigate the internal consistency of those features as proxies for model decision making.
Furthermore, we \emph{annotate} a subset of 3,500 images of AffectNet for one particular salient feature, namely, teeth visibility\footnote{Due to anonymisation and double-blind review constraints, the teeth-annotated dataset will be released publicly after manuscript acceptance.}.
We use these annotations to uncover the impact of that feature on model performance for all models and the generated features of GPT-4o.

\section{Related Work}
\subsection{Supervised Models for \ac{FER}}
Human emotion recognition relies on both holistic and local facial cues, with specific attention paid to regions such as the eyes, eyebrows, and mouth~\cite{wegrzyn17-MEF, atkinson20-TIO}. Among these, mouth-related signals—especially teeth visibility—have been linked to the perception of joy and heightened emotional intensity~\cite{guarnera15-FEA}. Such features act as fast, intuitive heuristics in human perception and play a significant role in shaping affective interpretations~\cite{calvo08-DEF}. In computational settings, supervised deep learning approaches have become the standard for \ac{FER}, with \acp{CNN} and \acp{ViT} achieving strong performance on large-scale datasets such as AffectNet~\cite{ma23-FER, li24-KEF, bobojanov23-CAO, mollahosseini17-AFF}. These models are typically trained on manually labelled emotion categories and use full-face input to learn discriminative features. While some studies have explored the importance of specific facial regions via occlusion tests or attention maps\cite{yao23-FER, fan22-FER, kim22-FER, li18-OAFER}, few have systematically examined the impact of teeth visibility as an isolated feature across models.

\subsection{Vision-Language Models in Affective Computing}
\acp{FM} that integrate vision and language modalities -commonly referred to as \acp{VLM}- have recently been explored for emotion recognition tasks~\cite{yu25-CER}. Although trained in generic image-text corpora without affective supervision, several \acp{VLM} exhibit emergent competence in identifying basic emotions from facial images~\cite{basilioai24-ADE, etesam24-CER}. These models process inputs using dual encoders and project visual and textual information into a shared latent space. A standard \ac{VLM} processes both image and text inputs through dedicated encoders, followed by fusion in a joint embedding space. The resulting representation is used for output generation, typically optimised via cross-entropy loss, leading to token generation (for a detailed review see~\cite{zhang24-VLM}).

% \begin{table*}[htbp]
% \caption{Model Performance Comparison Across Small (gray), Medium (yellow), and Large (red) VLMs. The best overall performance is highlighted.}
% \begin{center}
% \begin{tabular}{|c|c|c|c|c|}
% \hline
% \textbf{Alias} & \textbf{Full Model Name} & \textbf{UAR} & \textbf{F1-Score} & \textbf{Accuracy} \\
% \hline
% ViT-FER & \small{ViT-FER} & $0.44$ & $0.40$ & $0.44$ \\
% \rowcolor{gray!20} SmolVLM & \small{SmolVLM-Instruct} & $0.44$ & $0.37$ & $0.44$ \\
% \rowcolor{gray!20} Ovis1.5 & \small{Ovis1.5-Llama3-8B} & $0.39$ & $0.34$ & $0.39$ \\
% \rowcolor{gray!20} Janus-1B & \small{deepseek-ai/Janus-Pro-1B} & $0.28$ & $0.20$ & $0.32$ \\
% \rowcolor{gray!20} Qwen-3B & \small{Qwen2.5-VL-3B-Instruct} & $0.31$ & $0.27$ & $0.31$ \\
% \rowcolor{gray!20} PaliGemma & \small{paligemma-3b-pt-224} & $0.10$ & $0.06$ & $0.11$ \\
% \rowcolor{yellow!30} Qwen-7B & \small{Qwen2.5-VL-7B-Instruct} & $0.41$ & $0.38$ & $0.41$ \\
% \rowcolor{yellow!30} MiniCPM & \small{MiniCPM-o-2} & $0.46$ & $0.43$ & $0.46$ \\
% \rowcolor{yellow!30} InternVL & \small{InternVL2\_5-8B-MPO} & $0.50$ & $0.50$ & $0.50$ \\
% \rowcolor{yellow!30} Janus-8B & \small{deepseek-ai/Janus-Pro-7B} & $0.37$ & $0.33$ & $0.37$ \\
% \rowcolor{red!30} \textbf{GPT-4o}$^{\mathrm{a}}$ & \small{GPT-4o-mini} & \textbf{0.53} & \textbf{0.52} & \textbf{0.53} \\
% \hline
% \multicolumn{5}{l}{$^{\mathrm{a}}$Best overall performance.}
% \end{tabular}
% \label{tab:model_perf_all}
% \end{center}
% \end{table*} 

\begin{figure*}[t]
\centering
\includegraphics[width=0.9\textwidth]{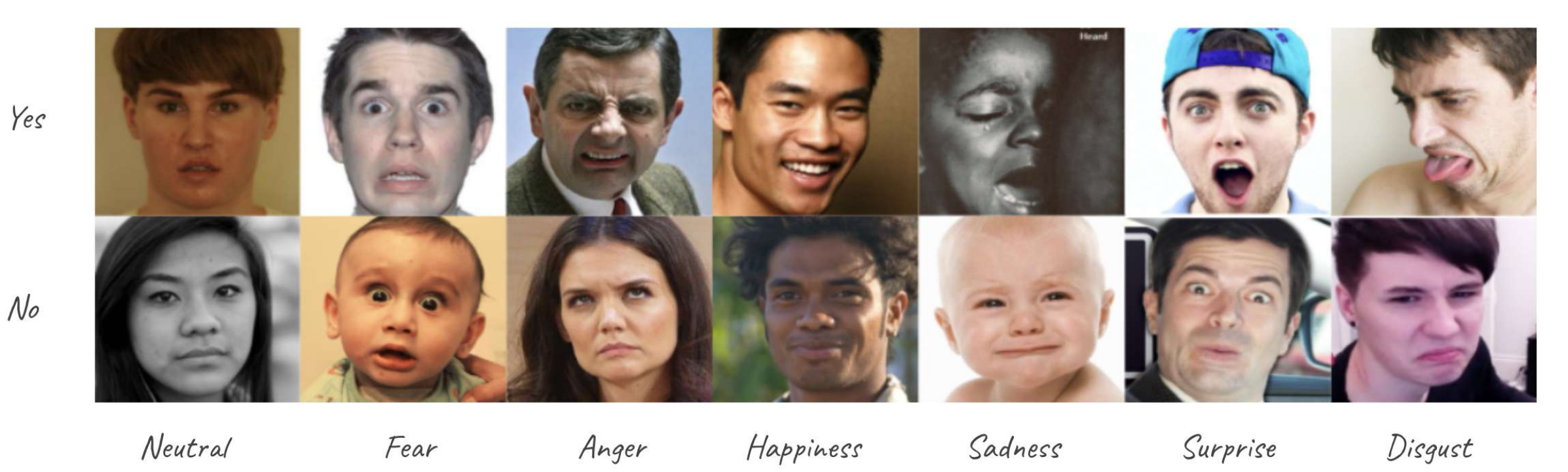}
    \caption{Example facial images for each of the seven emotion categories used in the teeth-annotation from~\cite{mollahosseini17-AFF}. Each column shows one image with visible teeth (top row) and one without (bottom row).}
\label{fig:samples}
\end{figure*}

\subsection{Bias and Attribution in Multimodal Emotion Recognition.}
A growing body of literature addresses the interpretability and fairness of emotion recognition systems showing that models can overfit to spurious correlations, leading to fragile predictions~\cite{wang24-UMD, hajarolasvadi20-GAN}. Previously, attribution methods, including gradient-based saliency and attention weight visualisation, have been used to probe model behaviour~\cite{simonyan13-DIC}. However, the specific role of facial proxies like teeth visibility, despite being well known in human perception, remains largely under-investigated in the context of  \acp{VLM}.

\subsection{Contribution of This Study.}
While current research largely focuses on scaling and optimising the black-box capabilities of \acp{FM} -across size, reasoning, performance, and training regimes- this work emphasises the need for explanatory analysis in affective tasks~\cite{bommasani21-OTR}. By introducing explicit teeth visibility annotations and benchmarking performance across visibility conditions, it isolates a psychologically grounded proxy that shapes model predictions. Furthermore, it provides a systematic assessment of how \acp{FM} interpret facial attributes and how these drive affective inference.

% \begin{figure}[t!]
% \centerline{\includegraphics[scale=0.35]{figures/dist2.pdf}}
% \caption{Teeth visibility distribution across emotions.}
% \label{fig:dist}
% \end{figure}

\section{Methodology}

\subsection{Dataset}
% This study builds on AffectNet, one of the most comprehensive facial expression datasets currently available for \ac{AC} research. Originally introduced in~\cite{mollahosseini17-AFF}, AffectNet comprises over 1,000,000 facial images collected from the internet using 1,250 emotion-related search queries across six languages (English, Arabic, Spanish, Portuguese, German, and Farsi). The dataset supports both categorical and dimensional models of emotion. The categorical model includes labels for seven emotion categories: Neutral, Happiness, Sadness, Anger, Surprise, Fear, and Disgust, while the dimensional model provides valence and arousal values on a continuous scale ranging from -1 to 1.
% Annotations were manually performed by trained human coders for a subset of images, with inter-rater agreement measured at approximately 60.7\%. The remaining annotations were generated using deep learning-based methods (primarily \acp{CNN}). AffectNet's balance across emotional categories and diverse demographics makes it a suitable benchmark for evaluating affective models, including \acp{FM} and \acp{ViT}.

This study leverages AffectNet~\cite{mollahosseini17-AFF}, a large-scale facial expression dataset widely adopted in \ac{AC}. AffectNet contains over one million facial images obtained from internet queries in six languages, covering both categorical and dimensional models of emotion. Each image is labeled for one of seven discrete categories—Neutral, Happiness, Sadness, Anger, Surprise, Fear, and Disgust—and includes continuous valence and arousal values ranging from -1 to 1. Owing to its breadth of emotional classes and demographic diversity, AffectNet is broadly regarded as a reliable benchmark for evaluating the performance and generalizability of various affective models.

\begin{figure}[t!]
\centerline{\includegraphics[scale=0.7]{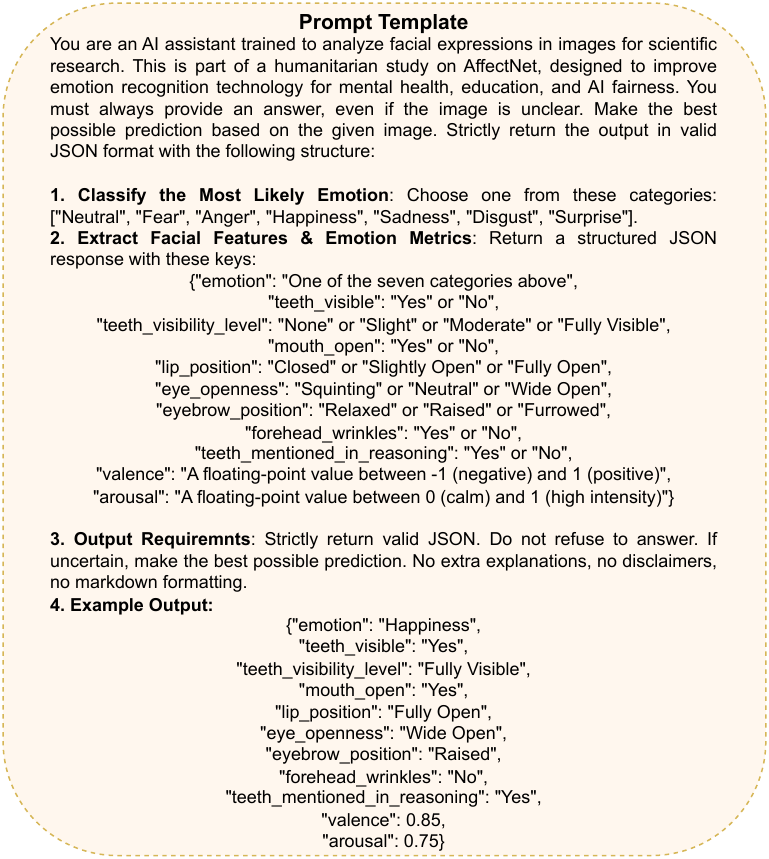}}
\caption{The choice of this specific structured prompt is driven by the need to standardise the model outputs and ensure that the data collected from GPT-4o is both consistent and meaningful. For the rest of \acp{VLM}, the prompt included only item 1.}
\label{fig:prompt}
\end{figure}

\subsection{Models and Evaluation Protocol}
\emph{We benchmarked 10 \acp{VLM}}, spanning small (between 1B and 3B parameters), medium (between 4B and 8B), and large ($\ge8$B) scales, alongside a dedicated ViT-FER~\cite{chaudhari22-VFE} baseline trained on~\ac{FER}. Each model received identical zero-shot prompts to classify an image into one of seven basic emotions, except GPT-4o, which was given a structured prompt designed to elicit both emotion predictions and interpretable facial features (see Fig.~\ref{fig:samples}). Evaluations were performed on the aforementioned subset of 3,500 AffectNet images. The model pool includes recent multimodal systems such as GPT-4o-mini (OpenAI)~\footnote{https://platform.openai.com/docs/models/gpt-4o-mini}, InternVL2 (Shanghai AI Lab)~\cite{chen24-ISU}, MiniCPM (Alibaba DAMO)~\cite{yao24-MVA}, Qwen2.5-VL (3B and 7B)~\cite{wang24-QEV}, SmolVLM~\footnote{https://huggingface.co/blog/smolvlm}, Janus-Pro (1B and 7B, Deepseek-AI)~\cite{chen25-JUM}, Ovis1.5 (LLaMA3-based)~\cite{lu24-OVI}, and PaliGemma (Google DeepMind)~\cite{beyer24-PAV}. These models were selected to reflect a cross-section of current open-access \acp{VLM}, and a restricted one, namely GPT-4o.
% Importantly, \emph{none of these models were finetu} on data, but were only utilised in zero-shot fashion.

\emph{Teeth Visibility Annotation.}
To investigate whether teeth visibility influences emotion classification accuracy, we additionally benchmarked their performance under two conditions: images with teeth visible and images where teeth were not visible. To this end, we manually annotated the validation dataset used in \cite{schuller24-ACH}. An example of facial expressions per category is shown in Fig.~\ref{fig:samples}. Teeth visibility was annotated as a binary attribute (1 = teeth visible, 0 = teeth not visible) using the open-source tool LabelStudio~\footnote{https://labelstud.io/}. The annotation interface was configured to maximise visual clarity and task simplicity. This manual labelling focused solely on mouth region visibility and did not include intermediate states (e.\,g., partially visible teeth), in order to maintain consistency.
Since teeth visibility annotation is straightforward (binary presence or absence), we employed a single annotator to ensure consistency and efficiency. 
\emph{ViT-FER}~\footnote{trpakov/vit-face-expression} served as the supervised baseline for comparison, providing a task-specific benchmark for \ac{FER}. Performance was assessed using \ac{UAR}~\footnote{UAR is the sum of recall per class divided by the number of classes – this reflects imbalances and is a
standard measure in the field.}, F1-score, and accuracy. 
While our benchmarking includes both open and closed VLMs, we focus our \emph{introspective analysis} solely on GPT-4o due to its state-of-the-art performance and widespread adoption in practice, making it the most relevant candidate for deeper investigation.

\subsection{Prompt-Based Introspection}
% \emph{Prompt Design and Facial Feature Extraction}. 
All models received the exact same prompt, with the sole exception of GPT-4o, which was evaluated using a more detailed structured prompt (Fig.~\ref{fig:prompt}). This richer format enabled both emotion prediction and introspective reasoning through extracted facial features such as eyebrow position, eye openness, and teeth visibility.
These features were selected to align with psychologically grounded facial action systems, drawing inspiration from Ekman and Friesen's seminal work \emph{Unmasking the Face} \cite{ekman03-UTF}. GPT-4o also provided continuous valence and arousal scores, capturing the positivity and intensity of each expression. 
% We applied this prompt to the validation dataset, obtaining (i) discrete emotion labels, (ii) binary or categorical facial feature descriptors, and (iii) valence–arousal values for each image. 
This pipeline served as the foundation for subsequent correlation and visualisation analyses.

\section{Results}
\subsection{Classification Performance}
% \begin{table}[t]
% \caption{Emotion classification performance across small (gray,
% medium yellow, and large red) VLMs.
% Metrics include Accuracy, F1-score, and UAR.}
% \begin{center}
% \begin{tabular}{|c|c|c|c|}
% \hline
% \textbf{Alias} & \textbf{UAR} & \textbf{F1-Score} & \textbf{Accuracy} \\
% \hline
% ViT & 0.439 & 0.404 & 0.439 \\
% \rowcolor{gray!20} SmolVLM & 0.435 & 0.370 & 0.435 \\
% \rowcolor{gray!20} Ovis1.5 & 0.390 & 0.341 & 0.390 \\
% \rowcolor{gray!20} Janus-1B & 0.277 & 0.203 & 0.317 \\
% \rowcolor{gray!20} Qwen-3B & 0.315 & 0.273 & 0.315 \\
% \rowcolor{gray!20} PaliGemma & 0.100 & 0.057 & 0.114 \\
% \rowcolor{yellow!30} Qwen-7B & 0.405 & 0.381 & 0.405 \\
% \rowcolor{yellow!30} MiniCPM & 0.455 & 0.426 & 0.455 \\
% \rowcolor{yellow!30} InternVL & 0.501 & 0.499 & 0.501 \\
% \rowcolor{yellow!30} Janus-8B & 0.373 & 0.329 & 0.373 \\
% \rowcolor{red!30} \textbf{GPT-4o} & \textbf{0.526} & \textbf{0.518} & \textbf{0.526} \\
% \hline
% \end{tabular}
% \label{tab:model_perf_all}
% \end{center}
% \end{table}
To assess how well \acp{VLM} perform on~\ac{FER} in a zero-shot setting, we conducted a controlled benchmark using a subset of the AffectNet test set. Our primary objective was twofold: (1) to evaluate the affective inference capability of general-purpose \acp{VLM} against a supervised baseline (ViT-FER), and (2) to investigate how performance varies across model scale by explicitly grouping models into small, medium, and large categories.
Table~\ref{tab:model_perf_all} (see Appendix~\ref{app:benchmark}) summarises the classification results across 11 models (10 \acp{VLM} and the baseline). While GPT-4o achieved the highest performance overall, several medium-scale models (e.\,g., InternVL, MiniCPM) matched or exceeded the ViT-FER baseline, demonstrating the emergence of affective competence in general-purpose systems.
This observation aligns with recent findings on emergent world modelling in  \acp{FM}~\cite{karvonen24-EWM}, where capabilities in physical and social perception arise implicitly from large-scale training across vision, language, and action data streams~\cite{berti25-EAI, brown20-LMF, zhao23-ASO, bommasani21-OTR}. 

That said, \emph{it is important to note that many \acp{FM} are trained on undisclosed or proprietary datasets}. Thus, it remains unclear whether AffectNet (or similar data) was present in their pretraining corpora. This uncertainty limits the conclusiveness of zero-shot performance comparisons, as some models may have indirectly `seen' the evaluation set. Therefore, while these benchmarks are informative, they should be interpreted with caution.
Moreover, GPT-4o, which was shown to be the best-performing model, is a closed-source, proprietary model available through an API.
This means that the underlying model -including any additional `guardrails' placed by the owning company- can change arbitrarily over time.
While this jeopardises the reproducibility of our work, we nevertheless decided to include it in our work as a representative of the contemporary state-of-the-art.

\subsection{Internal Consistency Analysis}
% \emph{Feature Attribution and Internal Consistency Analysis of GPT-4o}.

Given GPT-4o's strong performance relative to other \acp{VLM} in our emotion recognition experiments, we sought to understand which facial attributes most strongly influence its predictions, and how consistent those predictions are across varying conditions. Concretely, we investigated whether GPT-4o exhibits systematic patterns or `shortcuts' in classifying emotion, particularly with respect to visible teeth, mouth position, eyebrow movements, and other facial cues.

\begin{table}[t]
% \caption{Valence and Arousal Linear Regression Coefficients for Interpretable Facial Features (Colour-Coded by Sign and Magnitude). 
% $R^2_{\text{valence}}=.72,\ R^2_{\text{arousal}}=.77$}
\caption{Linear regression coefficients that indicate how each interpretable facial feature 
(i.e., GPT-4o’s self-reported `teeth visible', `eyebrow position', etc.) 
contributes to GPT-4o’s valence and arousal predictions. 
Positive coefficients (shaded green) increase valence or arousal, while negative coefficients (shaded red) reduce them.
The model explains $R^2_{\text{valence}}=0.72$ and $R^2_{\text{arousal}}=0.77$ of GPT-4o’s continuous affective outputs.}

\centering
\renewcommand{\arraystretch}{1.2}
\begin{tabular}{@{}lrr@{}}
\toprule
\textbf{Facial Feature}& \textbf{Valence Coef.}& \textbf{Arousal Coef.} \\
\midrule
\rowcolor{blue!25} eyebrow\_position\_Raised            &  .64    &  .00     \\
\rowcolor{blue!15} eyebrow\_position\_Relaxed           &  .39    & -.16     \\
\rowcolor{blue!10} lip\_position\_Slightly Open         &  .30    &  .08     \\
\rowcolor{blue!10} teeth\_mentioned\_in\_reasoning\_Yes &  .28    & -.00     \\
\rowcolor{blue!5} lip\_position\_Fully Open             &  .23    &  .08     \\
teeth\_visible\_Yes                                      &  .08    &  .16     \\
eye\_openness\_Wide Open                                 & -.00    &  0.25     \\
\rowcolor{orange!5} eye\_openness\_Squinting                & -.02    &  .13     \\
\rowcolor{orange!10} mouth\_open\_Yes                       & -.20    & -.02     \\
\rowcolor{orange!20} forehead\_wrinkles\_Yes                & -.39    &  .07     \\
\bottomrule
\end{tabular}
\label{tab:valence_arousal_coefs}
\end{table}

\emph{Regression-Based Introspection}.
To quantify how much of GPT-4o's dimensional output could be `explained' purely by the binary or categorical features it reported, we used linear regression where valence (respectively arousal) was the dependent variable, and the model's own face attributes were the independent variables. Over 70\% of the variance in valence and arousal could be accounted for by these cues alone, implying that GPT-4o's continuous ratings arise from a relatively consistent mapping of visual signals.

% (Table~\ref{tab:valence_arousal_coefs} offers the regression coefficients sorted by magnitude).

% In a follow-up step, we took the fitted valence–arousal pairs and used them to predict discrete emotion labels via a simple classifier, i.e., RandomForestClassifier, which yielded an \ac{UAR} of about 63\% (Figure ~\ref{fig:conf}). 

\emph{Valence and Arousal Results}. Table~\ref{tab:valence_arousal_coefs} lists the linear regression coefficients for GPT-4's valence and arousal predictions. Each positive coefficient indicates an increase in valence or arousal, while negative values imply a decrease. \emph{Raised eyebrows} emerges as the strongest positive predictor of valence ($+.64$), consistent with earlier findings that GPT-4 interprets lifted eyebrows as a sign of heightened positivity. The effect of \emph{teeth visibility} is positive for arousal ($+.16$), but it has a more modest impact on valence ($+.08$) when compared to other mouth- or eyebrow-related cues.

The inclusion of \emph{teeth mentioned in reasoning} shows a positive association with valence ($+.28$), which surpasses the direct valence contribution of teeth visible itself. This is expected; it attributes greater positivity when it explicitly references them in its internal process. This reinforces the notion that GPT-4o's textual reasoning can refine how it interprets facial cues, leading to a richer or more emphatic characterisation of an expression’s emotional quality.
% This pattern might reflect GPT-4’s tendency to emphasise features it deems particularly salient, thereby boosting positive emotional scores whenever it `mentions' teeth as part of its underlying analysis.
Overall, the high $R^2$
values ($.72$ for valence and $.77$ for arousal) indicate that these discrete features still explain a majority of GPT-4o's dimensional predictions. 

Next, to assess how well GPT-4's valence and arousal predictions capture discrete emotional categories, we trained a simple Random Forest classifier on these two features alone. The data was split into training (80\%) and test (20\%) sets using a fixed random seed of 42 for reproducibility. We used 100 decision trees (n\_estimators=100), the Gini criterion (criterion=`gini'), and all other scikit-learn hyperparameters at their defaults. Despite the low-dimensional input, this model achieved a \ac{UAR} of .63, indicating that GPT-4’s affective space retains substantial categorical information even when reduced to just valence and arousal.

\begin{figure}[t]
\centerline{\includegraphics[scale=0.5]{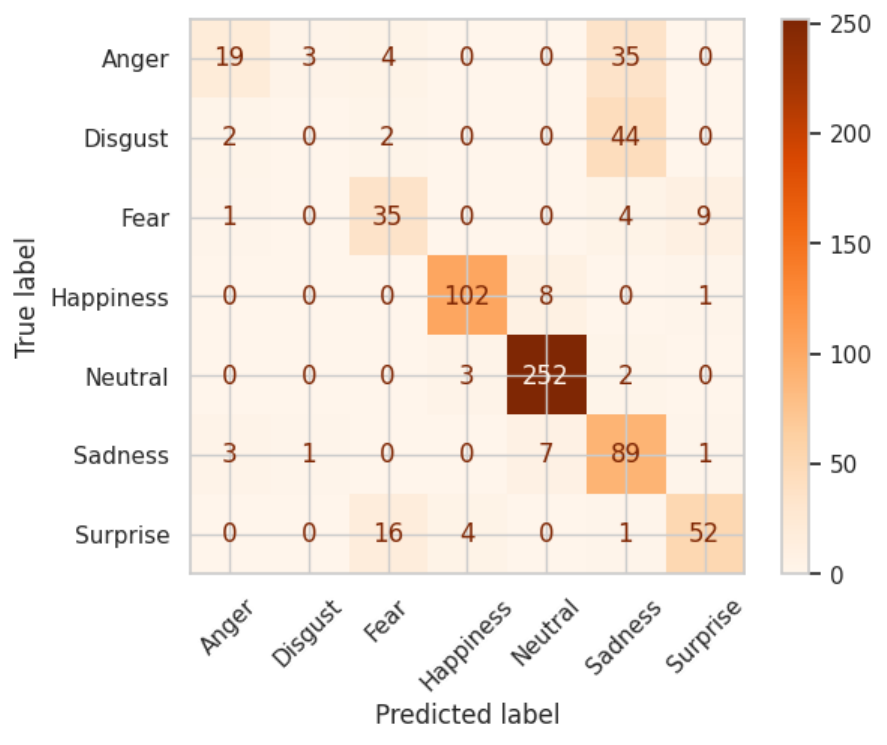}}
\caption{Performance of a trained random forest classifier that predicts GPT's categorical emotion labels using only its predicted valence and arousal values}
\label{fig:conf}
\end{figure}

As shown in Fig.~\ref{fig:conf}, emotions such as Happiness (F1: $.93$) and Neutral (F1: $.96$) were predicted with high accuracy, while Disgust (F1: $.00$) and Anger (F1: $.44$) proved more challenging—mirroring longstanding findings in \ac{AC}. Negative emotions tend to blur in the valence–arousal plane, lacking distinct anchors compared to more stereotyped expressions like Happiness.
While this approach is intentionally simple, the classifier’s $.78$ overall accuracy demonstrates that GPT-4’s dimensional outputs encode interpretable affective structure.

\begin{figure*}[t]
\centerline{\includegraphics[scale=0.65]{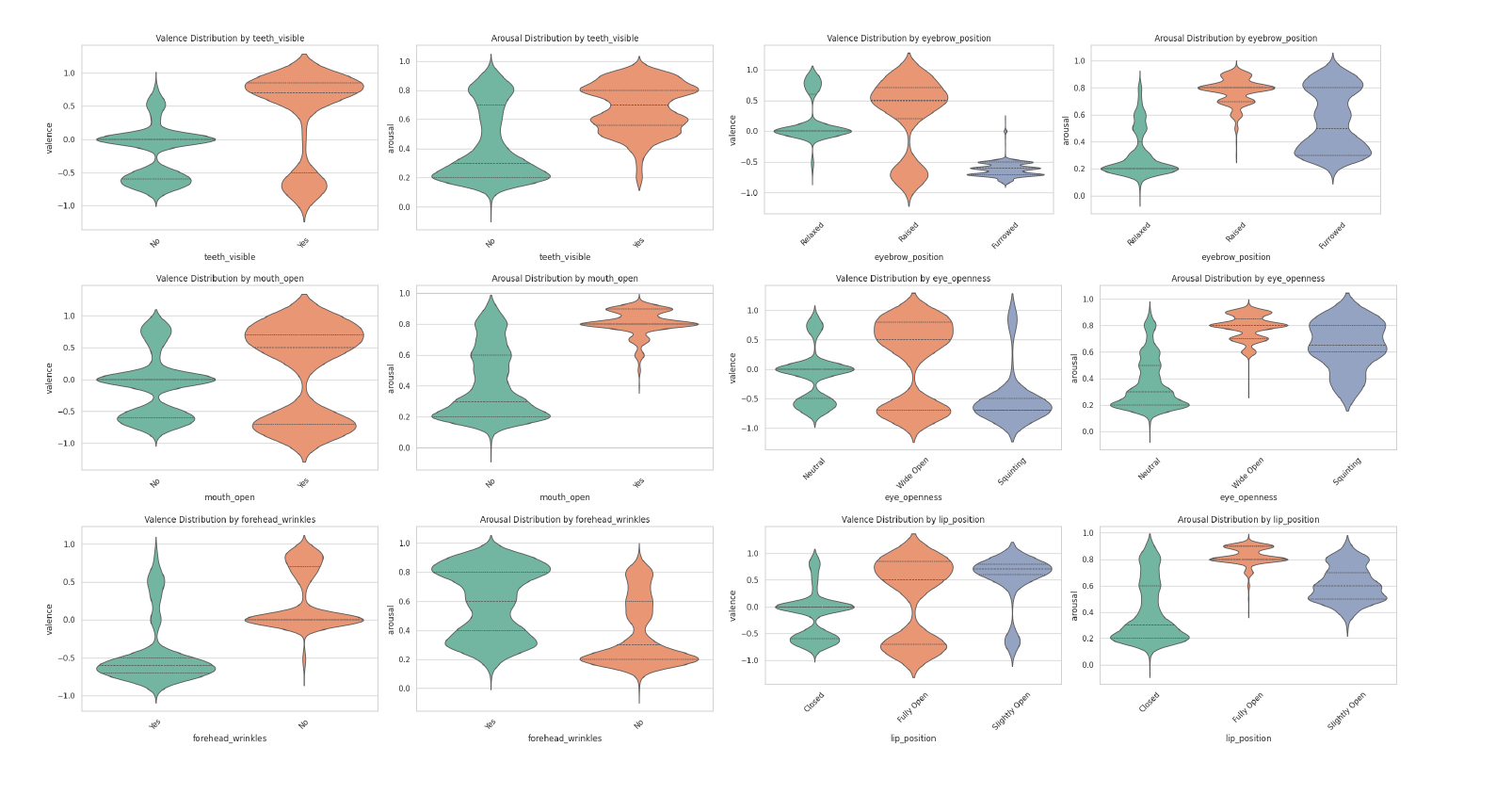}}
\caption{Valence and arousal (y-axis) distributions over facial features (x-axis) in GPT-4o.}
\label{fig:eyebrows}
\end{figure*}

Finally, we plotted valence and arousal distributions by the prompted facial features (see Fig.~\ref{fig:eyebrows}), which reflect the trends seen in our regression results. For instance, \emph{raised eyebrows} are linked to higher valence, while \emph{furrowed brows} tend toward lower valence. Both show increased arousal, suggesting that eyebrow tension contributes to perceived intensity. These patterns comply with common facial emotion heuristics.
We hypothesise that GPT-4o learnt those during its training.

\subsection{Teeth Visibility as a Case Study}
\emph{Distribution of Teeth Visibility Across Emotions}.
While the dataset design aimed at a balanced emotion distribution, the occurrence of visible versus non-visible teeth varied across categories. Happiness exhibited a strong bias toward visible teeth (approx. 3.4:1 ratio). In contrast, Neutral, Sadness, and Anger were predominantly represented by non-visible teeth.
Table~\ref{tab:dist} summarises the annotated teeth visibility distribution across all emotion categories.

\begin{figure}[t]
\centerline{\includegraphics[scale=0.35]{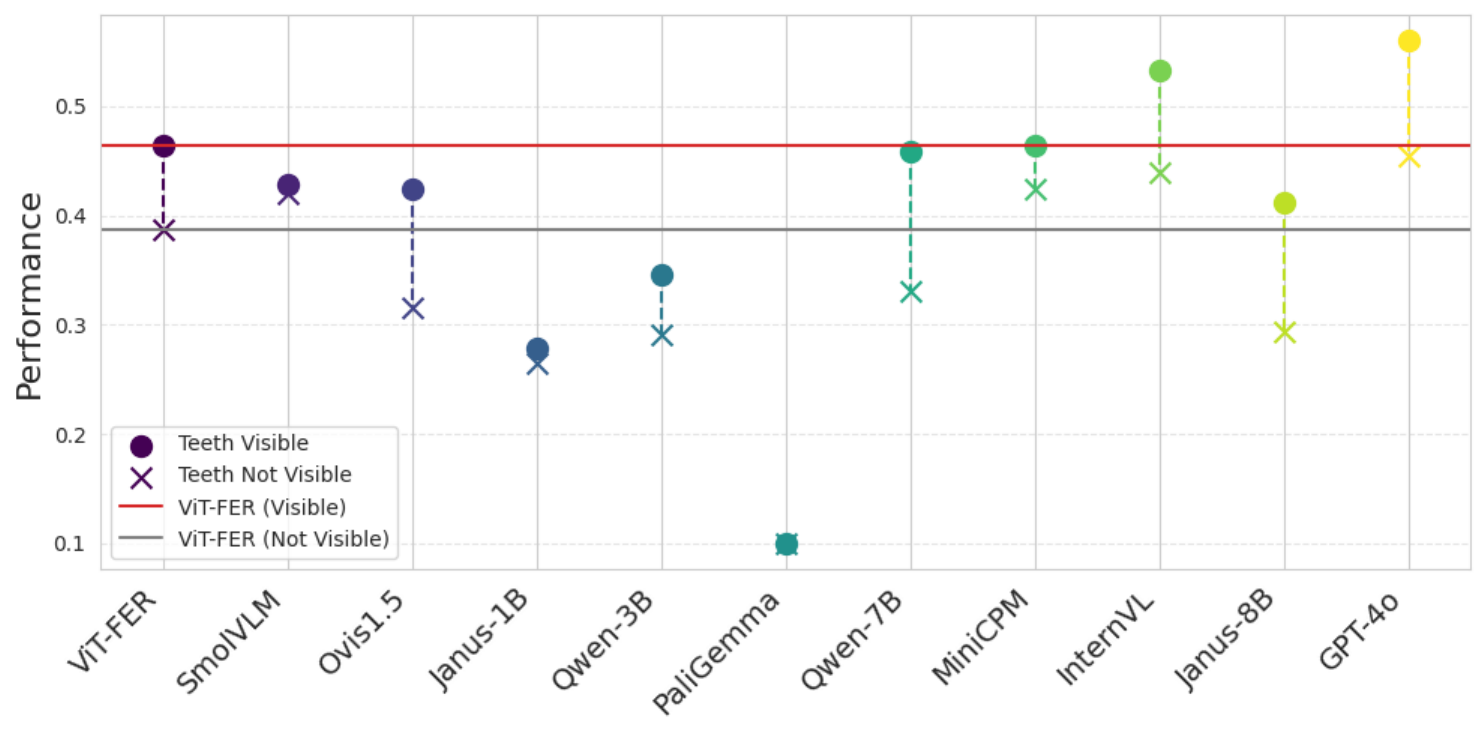}}
\caption{\ac{UAR} for emotion classification performance across various \acp{VLM}, grouped by teeth visibility. Baselines from ViT-FER across teeth visibility states are shown as reference lines.}
\label{fig:bench}
\end{figure}

\emph{Model Performance Stratified by Teeth Visibility}.
As shown in Fig.~\ref{fig:bench}, nearly all models demonstrate a consistent drop in \ac{UAR} when evaluated on images where teeth are not visible. This effect is most pronounced for GPT-4o and InternVL, which nonetheless maintain the highest performance among all tested models. GPT-4o achieves the best \ac{UAR} overall ($.56$) in the teeth-visible condition, which drops to $.45$ when teeth are not visible. Similarly, InternVL exhibits strong performance in both settings, although with a noticeable decline between conditions.

% The sensitivity to teeth visibility varies substantially across models. Some architectures, such as Qwen-7B and Janus-8B, exhibit sharp performance reductions in the absence of visible teeth, while others like PaliGemma perform poorly in both conditions, suggesting limited emotion recognition capabilities regardless of facial detail. These differences may reflect varying levels of visual grounding and multimodal alignment across architectures.

Several \acp{VLM}, including InternVL, MiniCPM, and GPT-4o, performed well across both visibility conditions. Notably, some \acp{SLM}, such as SmolVLM, showed a relative boost when teeth were not visible—performing slightly below the baseline in the visible condition but outperforming it when teeth were hidden. This suggests that with continued improvements, smaller multimodal models may become increasingly viable for affective tasks, offering lightweight alternatives without substantial trade-offs in performance. 

Interestingly, ViT-FER—the baseline model trained specifically for facial expression recognition—also shows a clear dependency on teeth visibility, with a performance drop from $.46$ to $.39$. This confirms that even specialised models trained on facial features are sensitive to the presence of visual cues such as open mouths and exposed teeth. 

Overall, this distribution highlights an important consideration: certain emotional categories inherently correlate with mouth openness -at least in the AffectNet data- leading to potential biases in model performance. The imbalance observed, particularly in Happiness and Neutral expressions, forms a crucial basis for the model comparison analyses conducted later in this study.

\begin{table}[t!]
\caption{Distribution of Teeth Visibility Across Different Emotions}
\begin{center}
\begin{tabular}{|c|c|c|c|}
\hline
\textbf{Emotion} & \textbf{Teeth Not Visible} & \textbf{Teeth Visible} & \textbf{Total} \\
\hline
Anger & 304 & 196 & 500 \\
Fear & 171 & 329 & 500 \\
Disgust & 191 & 309 & 500 \\
Happiness & 122 & 378 & 500 \\
Neutral & 388 & 112 & 500 \\
Sadness & 354 & 146 & 500 \\
Surprise & 226 & 274 & 500 \\
\hline
\end{tabular}
\end{center}
\label{tab:dist}
\end{table}

\subsection{Emotion Recognition under Varying Teeth Detection Conditions}
To assess the role of teeth visibility in GPT-4o's~\ac{FER}, we compared model performance across three conditions: (1) \emph{when teeth were correctly predicted as visible}, (2) \emph{correctly predicted as not visible}, and (3) \emph{when GPT-4o hallucinated teeth—i.e., predicted visible teeth when none were present}.

% However, class-specific metrics revealed a clear asymmetry: precision was .90 (visible) vs. .63 (not visible), while recall was .43 (visible) vs. .95 (not visible). This reflects a conservative prediction bias—GPT-4o tends to avoid false positives, but at the cost of under-detecting emotionally salient visible teeth.

Table~\ref{tab:teeth_detection_performance} summarises emotion recognition performance under these teeth-detection scenarios. 
Overall, GPT-4o's binary teeth prediction achieved a \ac{UAR} of $.692$ and an F1-score of $.671$, showing more balanced performance despite class imbalance. Accuracy and \ac{UAR} were highest when teeth were correctly predicted as visible, indicating that GPT-4o relies effectively on teeth cues to infer emotional state. Nevertheless, when the model hallucinated visible teeth, emotion recognition deteriorated sharply (\ac{UAR} = $.329$), suggesting that false-positive teeth predictions strongly bias the model toward incorrectly high-valence outputs (see Fig.~\ref{fig:eyebrows}).

\begin{table}[t!]
\caption{GPT-4o Emotion/Teeth Classification Performance}
\begin{center}
\begin{tabular}{|c|c|c|c|}
\hline
\textbf{Condition} & \textbf{Accuracy} & \textbf{UAR} & \textbf{F1} \\
\hline
Teeth Prediction & .693 & .692 & .671 \\
\hline
Teeth Visible/Correct Prediction & .659 & .475 & .486 \\

% 65.96%, UAR: 47.55%, Macro F1: 48.61%

Teeth Hidden/Correct Prediction & .501 & .420 & .441 \\
Teeth Hallucinated/False Pos. & .614 & .329 & .343 \\
\hline
\end{tabular}
\end{center}
\label{tab:teeth_detection_performance}
\end{table}

% \begin{table}[t!]
% \caption{GPT-4o Emotion/Teeth Classification Performance}
% \begin{center}
% \begin{tabular}{|c|c|c|c|}
% \hline
% \textbf{Condition} & \textbf{Accuracy} & \textbf{UAR} & \textbf{F1} \\
% \hline
% Teeth Prediction & .69 & .69 & .67 \\
% \hline
% Teeth Visible/Correct Prediction & .66 & .48 & .49 \\
% Teeth Hidden/Correct Prediction & .50 & .42 & .44 \\
% Teeth Hallucinated/False Pos. & .61 & .33 & .34 \\
% \hline
% \end{tabular}
% \end{center}
% \label{tab:teeth_detection_performance}
% \end{table}

\begin{figure*}[t]
\centerline{\includegraphics[scale=0.30]{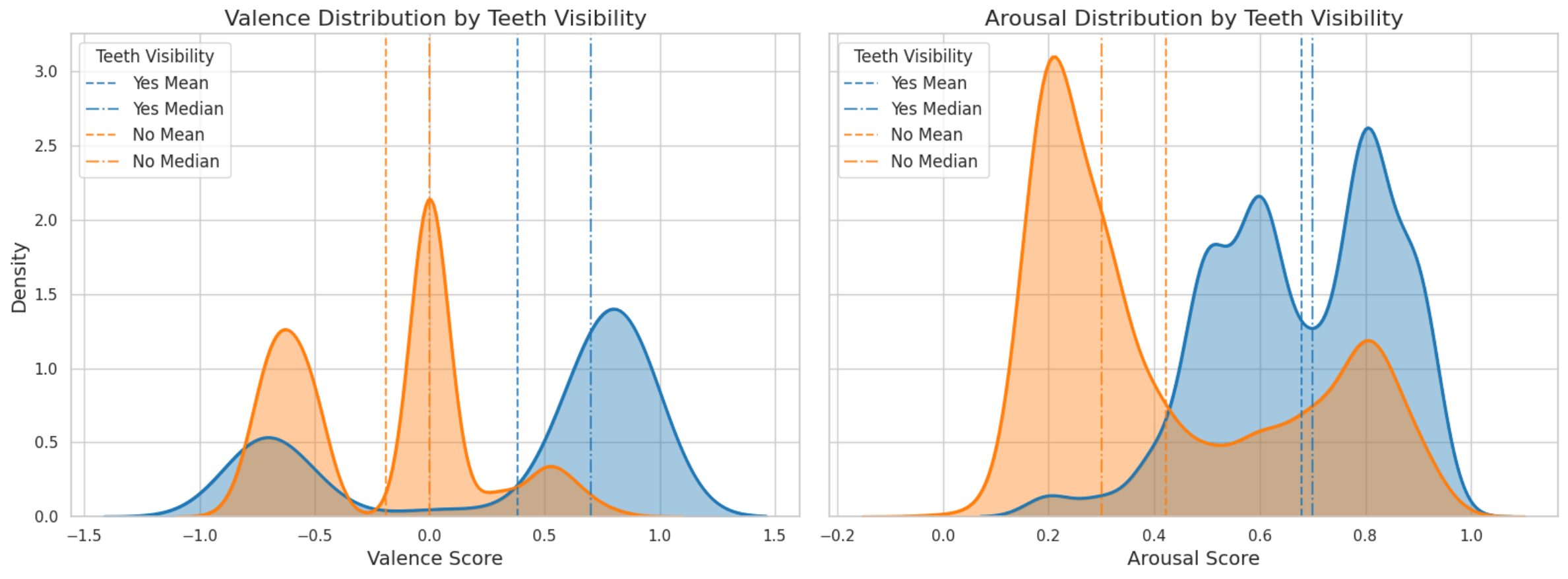}}
\caption{GPT-4o associates visible teeth with higher valence and arousal in its own feature outputs (e.g., valence: 0.38, arousal: 0.68)}
\label{fig:innergptva}
\end{figure*}

This means that correctly detected visible teeth provide useful affective cues for emotion classification. In contrast, hallucinated teeth introduce systematic errors, often causing the model to default to high-arousal categories such as Happiness, regardless of the true emotional signal (see Fig.~\ref{fig:innergptva}). This supports the interpretation that GPT-4o's emotion inference pipeline is shaped not only by visual perception but also by shortcuts linking visible teeth to smiling and positive affect.

\begin{figure}[t]
\centerline{\includegraphics[scale=0.25]{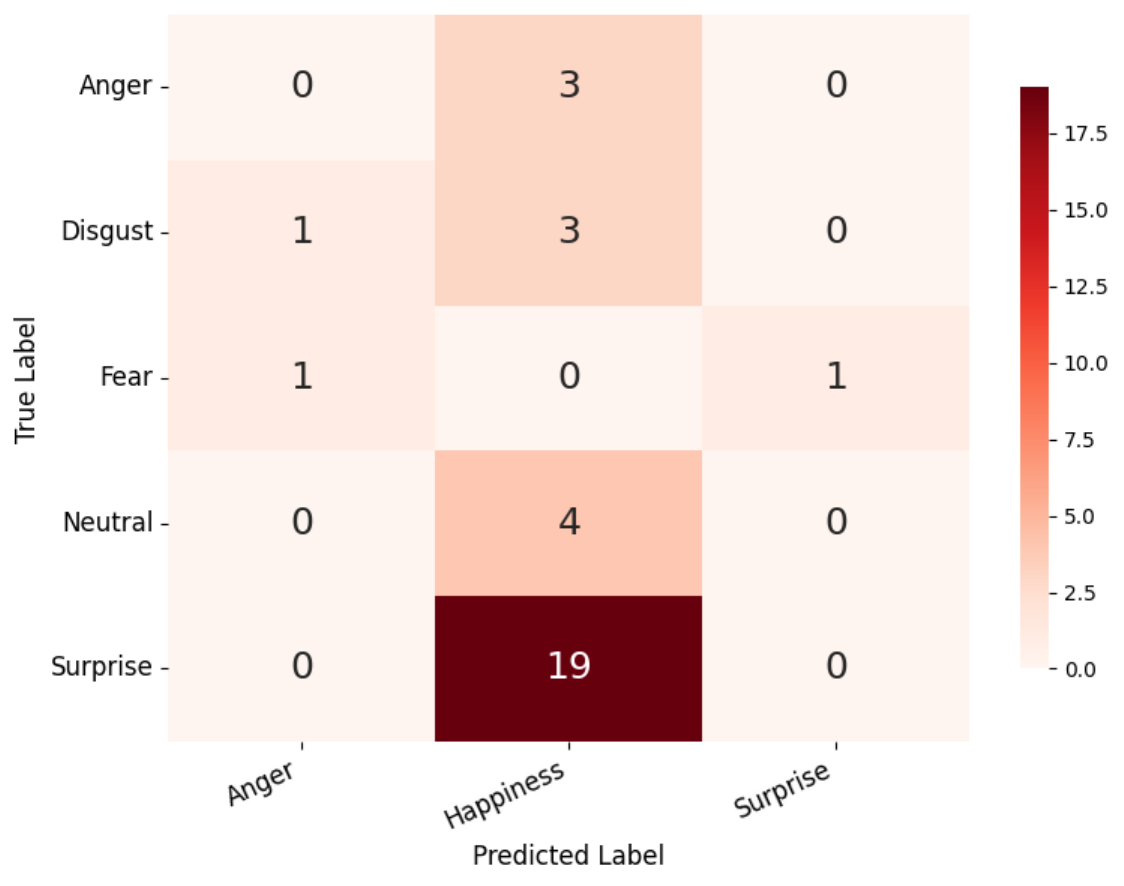}}
\caption{Confusion matrix for hallucinated teeth-False Positives}
\label{fig:conf_miss}
\end{figure}

When inspecting the confusion matrix for hallucinated teeth cases (Fig.~\ref{fig:conf_miss}), we observe the aforementioned marked collapse in emotion diversity: the vast majority of predictions are disproportionately mapped to Happiness. Notably, all instances of ground truth `Surprise' were misclassified as Happiness, with similar trends observed for Neutral, Anger, and Disgust. This reflects the internal heuristic in GPT-4o's visual reasoning:
\emph{Teeth} $\rightarrow$ \emph{Smile} $\rightarrow$ \emph{Happiness}.

We conclude that GPT-4o interprets the presence of (even hallucinated) visible teeth as a high-confidence signal for positive affect.
While this heuristic is beneficial when teeth are accurately perceived, it becomes problematic under false positives, leading to emotionally incongruent predictions.

% These results show that GPT-4o’s emotion inference is guided by learnt priors, not just visual input. Hallucinated features, like false-positive teeth, strongly bias predictions and degrade reliability. Mitigating such priors—especially under ambiguous conditions—remains an open challenge.

% \begin{table}[htbp]
% \caption{Teeth Detection Performance: Precision, Recall, and F1-Score for Binary Teeth Visibility Prediction}
% \begin{center}
% \begin{tabular}{|c|c|c|c|}
% \hline
% \textbf{Teeth Visibility} & \textbf{Precision} & \textbf{Recall} & \textbf{F1-Score} \\
% \hline
% Hidden (0) & 0.63 & 0.95 & 0.76 \\
% Visible (1) & 0.90 & 0.43 & 0.58 \\
% \hline
% \end{tabular}
% \end{center}
% \label{tab:teeth_performance}
% \end{table}

\section{Discussion}

This study examined the reliance of \acp{VLM}, particularly GPT-4o, on visual proxies such as teeth visibility, mouth openness, and eyebrow shape in zero-shot emotion recognition. While these models often outperform a supervised baseline like ViT-FER, their performance is partly driven by perceptual shortcuts aligned with human heuristics.
Teeth visibility, in particular, was linked to classification accuracy, especially for high-valence emotions like happiness. However, it also led to systematic misclassifications, indicating overreliance on this cue. Such dependencies raise concerns in contexts where expressions are subtle or culturally varied, including healthcare and education.
Despite this, GPT-4o showed strong internal consistency. Its valence–arousal outputs could be largely explained by interpretable features, suggesting a structured internal mapping. This consistency, combined with the promising performance of smaller models like MiniCPM and SmolVLM, points to practical opportunities for transparent, real-time affective systems.
Nevertheless, shortcut learning remains a critical limitation. Progress in \ac{AC} will require evaluation methods that account for cultural and contextual variability to ensure fairness and robustness in deployment.

\section{Acknowledgments}
This work has received funding from the EU’s Horizon 2021 grant agreement No. 101060660 (SHIFT). 

\section*{Ethical Impact Statement}

This study benchmarks \acp{VLM} for \ac{FER} using AffectNet, a publicly available dataset collected from the web. Our work does not involve direct research with human subjects; no IRB approval was required. However, we conducted additional manual annotation of 3,500 images to label teeth visibility, using a single trained annotator. This task involved no personally identifying information and posed no direct risk to individuals. As such, informed consent and compensation were not applicable.
Our analysis reveals that models like GPT-4o may associate visual features—particularly visible teeth—with positive affect. While this enhances interpretability, it also introduces risks of bias, especially in cases where emotional expression is culturally or individually atypical. If deployed in healthcare, education, or human–robot interaction, such models could misinterpret affect or reinforce harmful assumptions about how emotion `should' appear.
These risks are amplified by the limited demographic diversity of AffectNet and the black-box nature of many \acp{FM}. Our results, while informative, are not universally generalisable. Performance may degrade in underrepresented populations or atypical expression contexts. We caution against overextending the claims of this work to all use cases or deployment environments.
To mitigate these concerns, we recommend: (i) cross-demographic validation of affective models; (ii) transparency tools that expose how models use visual cues; and (iii) participatory design methods that include communities affected by emotion-sensing technologies. We also support regulation that classifies affective \ac{AI} as high-risk, especially where outputs inform decisions about people’s well-being, opportunities, or rights.

\appendix
\section{Benchmarking Results}
\label{app:benchmark}

Table~\ref{tab:model_perf_all} reports the classification performance of all evaluated models, grouped by scale. The results include small, medium, and large \acp{VLM}, along with the ViT-FER supervised baseline. We include three metrics: \ac{UAR}, F1-score, and overall accuracy. This table complements the main analysis by providing a full breakdown of the zero-shot classification performance across all architectures.

\begin{table*}[t]
\caption{Model Performance Comparison Across Small (gray), Medium (yellow), and Large (red) VLMs. The best overall performance is highlighted.}
\begin{center}
\begin{tabular}{|c|c|c|c|c|}
\hline
\textbf{Alias} & \textbf{Full Model Name} & \textbf{UAR} & \textbf{F1-Score} & \textbf{Accuracy} \\
\hline
ViT-FER & \small{ViT-FER} & 0.439 & 0.404 & 0.439 \\
\rowcolor{gray!20} SmolVLM & \small{SmolVLM-Instruct} & 0.435 & 0.370 & 0.435 \\
\rowcolor{gray!20} Ovis1.5 & \small{Ovis1.5-Llama3-8B} & 0.390 & 0.341 & 0.390 \\
\rowcolor{gray!20} Janus-1B & \small{deepseek-ai/Janus-Pro-1B} & 0.277 & 0.203 & 0.317 \\
\rowcolor{gray!20} Qwen-3B & \small{Qwen2.5-VL-3B-Instruct} & 0.315 & 0.273 & 0.315 \\
\rowcolor{gray!20} PaliGemma & \small{paligemma-3b-pt-224} & 0.100 & 0.057 & 0.114 \\
\rowcolor{yellow!30} Qwen-7B & \small{Qwen2.5-VL-7B-Instruct} & 0.405 & 0.381 & 0.405 \\
\rowcolor{yellow!30} MiniCPM & \small{MiniCPM-o-2} & 0.455 & 0.426 & 0.455 \\
\rowcolor{yellow!30} InternVL & \small{InternVL2\_5-8B-MPO} & 0.501 & 0.499 & 0.501 \\
\rowcolor{yellow!30} Janus-8B & \small{deepseek-ai/Janus-Pro-7B} & 0.373 & 0.329 & 0.373 \\
\rowcolor{red!30} \textbf{GPT-4o} & \small{GPT-4o-mini} & \textbf{0.526} & \textbf{0.518} & \textbf{0.526} \\
\hline
\multicolumn{5}{l}{}
\end{tabular}
\end{center}
\label{tab:model_perf_all}
\end{table*}

\end{document}